\documentclass[sigconf]{acmart}
\usepackage{booktabs} 
\usepackage{subfig}
\usepackage{amsmath}

\setcopyright{rightsretained}

\acmConference[KDD]{KDD Fashion Workshop}{August 2018}{London, United Kingdom}
\acmYear{2018}
\copyrightyear{2018}

\acmArticle{4}
\acmPrice{15.00}

\DeclareMathOperator*{\EX}{\mathbb{E}}
\DeclareMathOperator{\real}{\mathbf{x}}
\DeclareMathOperator{\fake}{\mathbf{\tilde{x}}}
\DeclareMathOperator{\vecfake}{\mathbf{\tilde{v}}}
\DeclareMathOperator{\col}{\mathbf{c}}
\DeclareMathOperator{\texture}{\mathbf{t}}
\DeclareMathOperator{\laplacian}{\mathbf{\tilde{S}}}
\DeclareMathOperator{\shape}{\mathbf{s}}
\DeclareMathOperator{\loss}{\mathcal{L}}
\DeclareMathOperator{\Tr}{Tr} 
\DeclareMathOperator*{\argmin}{argmin}
\newcommand{\Ltwo}[1]{\big|\big|#1\big|\big|_2^2}
\newcommand{\Lone}[1]{\big|\big|#1\big|\big|_1}

\begin{document}
\title{Disentangling Multiple Conditional Inputs in GANs}

\author{G\"{o}khan Yildirim}
\affiliation{
  \institution{Zalando Research}
  \city{Berlin}
  \state{Germany}
}
\email{gokhan.yildirim@zalando.de}

\author{Calvin Seward}
\affiliation{
	\institution{Zalando Research}
	\city{Berlin}
	\state{Germany}
}
\additionalaffiliation{
	\institution{LIT AI Lab \& Institute of Bioinformatics, Johannes Kepler University}
	\city{Linz}
	\state{Austria}
}
\email{calvin.seward@zalando.de}

\author{Urs Bergmann}
\affiliation{
	\institution{Zalando Research}
	\city{Berlin}
	\state{Germany}
}
\email{urs.bergmann@zalando.de}

\renewcommand{\shortauthors}{G. Yildirim et al.}

\begin{abstract}
In this paper, we propose a method that disentangles the effects of multiple input conditions in Generative Adversarial Networks (GANs). In particular, we demonstrate our method in controlling color, texture, and shape of a generated garment image for computer-aided fashion design. To disentangle the effect of input attributes, we customize conditional GANs with consistency loss functions. In our experiments, we tune one input at a time and show that we can guide our network to generate novel and realistic images of clothing articles. In addition, we present a fashion design process that estimates the input attributes of an existing garment and modifies them using our generator.

\end{abstract}


\maketitle

\section{Introduction}
The process of fashion design requires extensive amount of knowledge in creation and production of garments. Designers often need to closely follow the current trends and predict what will be popular in the future. Therefore, to be ahead of the curve for commercial success, a well-structured and agile design process is crucial. A machine-assisted design approach that combines human experience with deep learning can help designers to rapidly visualize an original garment and can save time on design iteration cycles. 

In recent years, deep learning techniques have been used in various fashion related topics, such as article representation and retrieval~\cite{deep_fashion, fashion_128, fashion_dna}. In addition, variants of Generative Adversarial Networks (GAN)~\cite{GAN} opened up image generation and manipulation possibilities that are, among other things, especially effective in fashion visualization and design~\cite{manipulation_manifold, image_to_image, fashion_style_generator, fashion_synthesis, cagan}.

GANs are a deep learning architecture that learns to map an easy-to-sample latent probability distribution into a complex and high-dimensional distribution, such as images. In its bare form, GANs do not provide out-of-the-box control for the generation process. Researchers have achieved control of image generation by using GANs that are conditioned on a categorical input~\cite{conditional_gan, ac_gan}. In this paper, we employ conditional GANs to control the visual attributes, such as color, texture, and shape, of a generated apparel.

One of the main challenges of the conditional image generation GANs is to isolate the effects of input attributes on the final image. For example, we want the color of an article to stay constant, when we tune its texture and/or shape. One possibility would be to employ Adversarial Autoencoders~\cite{AVAE} or DNA-GAN~\cite{dna_gan} to disentangle the inputs. However, this requires an exhaustive dataset, in other words, we need to have the images of garments with comprehensive color, texture, and shape combinations. Unfortunately, most of the clothing articles are produced in a limited range of design options.

In this paper, we introduce a conditional GAN architecture and a training procedure that independently controls the color, texture, and shape characteristics of a generated garment image. In order to disentangle the influence of generator inputs,  for each attribute, we add a consistency loss function at the output of our generator. We show that when we tune one attribute, the other two characteristics of the generated image stay visually stable. We then demonstrate a simple fashion-design process by first estimating the characteristics of a real article, and then modifying these properties using our generator inputs. Finally, we discuss possible future directions to add more control over the generation process.
\vspace{-0.1cm}
\begin{figure}[h]
	\centering
	\includegraphics[width=0.42\textwidth]{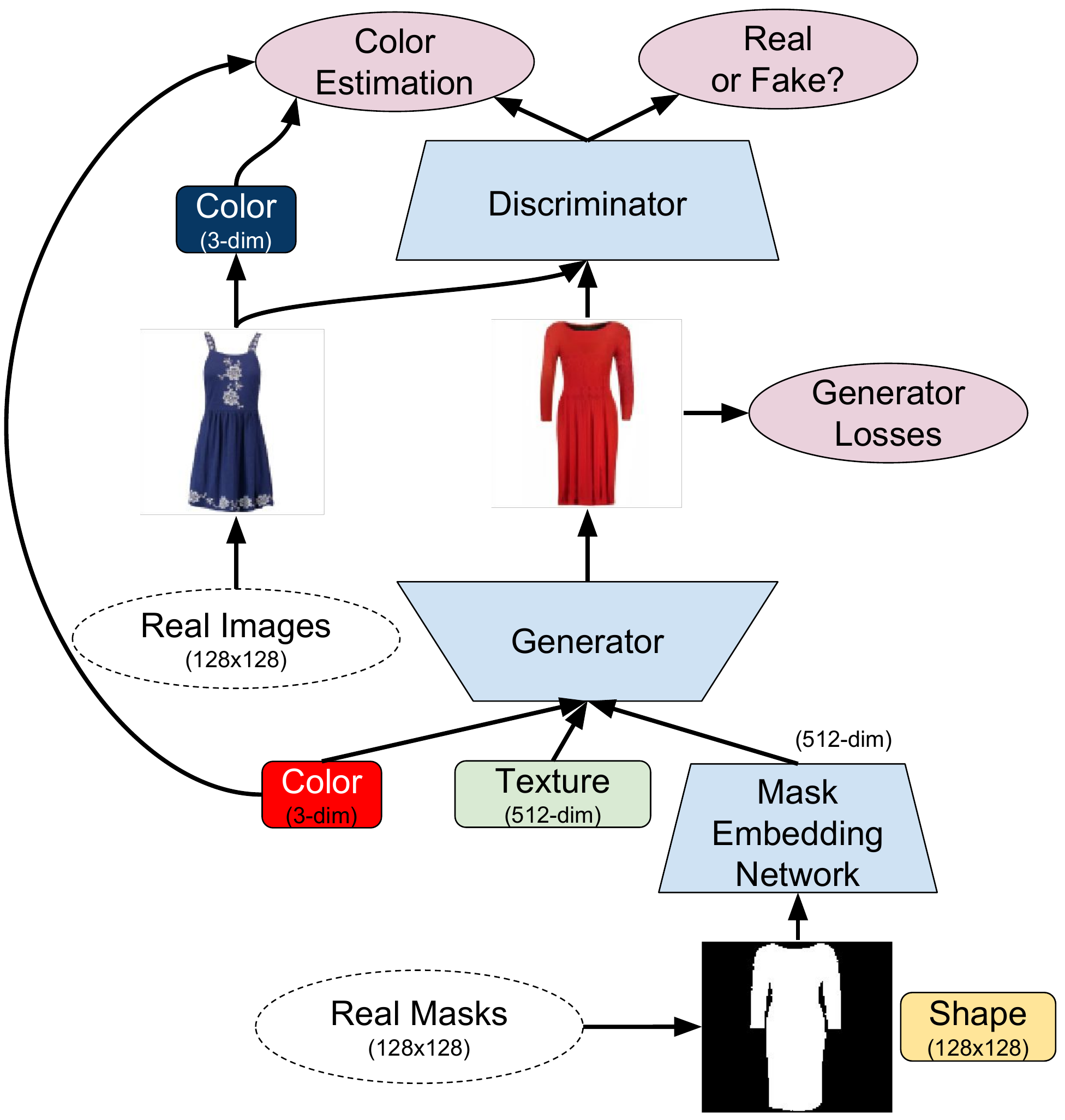}
	\vspace{-0.1cm}
	\caption{The flowchart of our method. A generated image, which is formed by using color, texture, and shape attributes, is compared against real article images by our discriminator.}
	\label{fig:architecture}
\end{figure}

\section{Method}
An overview of our method can be found in Figure~\ref{fig:architecture}. Our generator has three inputs. The first one represents the average color of an article, $\col\sim p(\col)$. The color is defined as a 3-dimensional vector of RGB values. We choose $p(\col)$ to be uniform between the interval [-1, 1], so that we can generate clothing articles with any desired color. The second input is a 512-dimensional latent vector that represents the texture and the local structure of an article $\texture\sim p(\texture)$, where $p(\texture)$ is a normal distribution with zero mean and unit variance. The final input, which represents the shape of an article, is a binary segmentation mask $\shape\sim p(\shape)$ of size $128 \times 128$ pixels. $p(\shape)$ is the distribution of segmentation masks of real articles. In order to feed the shape input, we use a mask embedding network that transforms the binary image into a 512-dimensional vector. Note that the mask embedding network is a part of the image synthesis and is jointly trained with our generator. All three attributes are concatenated into a 1027-dimensional vector and are passed into the generator, resulting in $\fake = G(\col, \texture, \shape)$, where $\fake$ is a generated image.

Our discriminator $D(\cdot)$ (or critic) outputs a 4-dimensional vector and is trained to perform two tasks: it distinguishes (in 1-dim) between real $\real$ and fake $\fake$ images and it estimates the average color (in 3-dim) of an input article. This is very similar to Auxiliary Classifier (AC) GANs~\cite{ac_gan}, except that we replace the categorical classification with a color regression.

In order to train our GAN architecture, we use the WGAN-GP loss~\cite{improved_wgan}. The first component of this loss represents a distance between real and generated image distributions, and is defined as follows:
\begin{equation}
\loss_{W} = \EX_{\real \sim P_r} \big[D'(\real)\big] - \EX_{\fake \sim P_g} \big[D'(\fake) \big],
\end{equation}
where $\loss_{W}$ is the Wasserstein loss in~\cite{improved_wgan}, $D'(\cdot)$ is the real-vs-fake output of our discriminator, $\real$ is a real image that comes from the distribution $P_r(\real)$ and $\fake$ is a generated image that comes from $P_g(\fake)$ and is induced by our generator $G(\cdot)$. As mentioned before, the auxiliary output of the discriminator attempts to correctly estimate the average color of real and generated article images, which is computed using the following function:
\begin{equation}
A(\real, \shape) = \frac{1}{|\shape|} \sum_{i,j}\shape(i,j) \cdot \real(i, j).
\label{eq:average-color}
\end{equation}
Here, $A(\cdot, \cdot)$ is a function that calculates the average color of a real or fake article using its corresponding segmentation mask. The sum is computed over image pixel locations $(i,j)$ and then normalized by $|\shape|$, which is the number of pixels inside the binary segmentation mask. From this point forward, we will drop the pixel locations $(i,j)$ from our notation. The auxiliary loss is defined as:
\begin{equation}
\loss_{\text{aux}} = \EX_{\substack{\real \sim P_r\\ \shape\sim p(\shape)}}\Big[\Ltwo{D^{\text{aux}}(\real) - A(\real, \shape)}\Big] + \EX_{\substack{\fake \sim P_g\\ \shape\sim p(\shape)}}\Big[\Ltwo{D^{\text{aux}}(\fake) - A(\fake, \shape)}\Big].
\end{equation}
Here, $\loss_{\text{aux}}$ is the auxiliary loss for color estimation, $D^{\text{aux}}(\cdot)$ is the auxiliary output of the discriminator, which is trained by optimizing the following function:
\begin{equation}
\max_{D} \loss_{W} - \loss_{\text{aux}} - \lambda_{\text{gp}}\loss_{\text{gp}},
\end{equation}
where, $\loss_{\text{gp}}$ is the gradient penalty term from~\cite{improved_wgan} and its weight is $\lambda_{\text{gp}} = 10$.

We have multiple generator inputs that influence the appearance of a synthesized image. In order to disentangle the effects of these inputs, we use three loss functions that works in collaboration and minimizes the crosstalk when we individually tune the attributes.

\subsection{Color Consistency}
We want to control the average color of the generated article through the 3-dimensional color input. Consider that we generate two images, namely $\fake_1^{\col} = G(\col, \texture_1, \shape_1)$ and $\fake_2^{\col}=G(\col, \texture_2, \shape_2)$, that are synthesized with the same color and different texture and shape inputs. We can consider these images as independent random variables that are identically distributed with $P_g^{\col}=P_g(\fake | \col)$. We can achieve color disentanglement at the output of our generator by enforcing the average color of the generated article images to be the same:
\begin{equation}
\loss_{c} = \EX_{\col\sim p(\col)} \bigg[\EX_{
	\substack{\fake_1^{\col}, \fake_2^{\col}\sim P_g^{\col} \\ \shape_1,\shape_2\sim p(\shape)}
}\Big[\Ltwo{A(\fake_1^{\col}, \shape_1) - A(\fake_2^{\col}, \shape_2\big)}\Big]\bigg],
\label{eq:color-consistency}
\end{equation}
where $\loss_{c}$ is defined as the color consistency loss. The inner expectation ensures that for a given color input $\col$, all the generated images have approximately the same average color. The outer expectation provides color consistency for all colors.

\subsection{Texture Consistency}
In order to provide texture consistency, we need to preserve the local structure (garment pattern, wrinkles, and shading), even when we change the color and shape inputs. In photo-realistic style transfer~\cite{deep_style}, to retain the details of an input image, Laplacian matting matrices~\cite{closed_matting} are employed. These very sparse matrices represent the local structure around each pixel and can be used to measure structural similarity between a source and a target image. In our method, to ensure texture consistency, we adopt a similar approach. Let $\fake_1^{\texture}$ and $\fake_2^{\texture}$ be two images that are generated with the same texture, but different colors and shapes. Similar to what we do in color consistency, we consider these images as independent random variables that are identically distributed with $P_g^{\texture}=P_g(\fake | \texture)$. Prior to loss computation, we define the following operations:
\begin{equation}
\vecfake_1^{\texture} = V(\fake_1^{\texture}), \quad \laplacian_1^{\texture} = S(\fake_1^{\texture}),
\end{equation}
where $V(\cdot)$ is an operator that flattens an input image into a $128^2 \times 3$ matrix and $S(\cdot)$ computes the Laplacian matting matrix ($128^2 \times 128^2$) described in~\cite{closed_matting}. In order to minimize texture inconsistencies, we propose the following loss function:
\begin{equation}
\loss_{t} = \EX_{\texture\sim p(\texture)} \bigg[\EX_{\fake_1^{\texture}, \fake_2^{\texture}\sim P_g^{\texture}} \Big[ \Tr\big({\vecfake_1^{\texture}}^T\laplacian_2^{\texture} \vecfake_1^{\texture} + {\vecfake_2^{\texture}}^T\laplacian_1^{\texture} \vecfake_2^{\texture}\big) \Big] \bigg],
\label{eq:texture-consistency}
\end{equation}
where $\loss_{t}$ is the texture consistency loss, and $\Tr(\cdot)$ is the trace of a matrix. In equation~\ref{eq:texture-consistency}, the inner expectation minimizes local structural differences between two images that are generated with the same texture input, regardless of their colors and shapes. The outer expectation ensures that consistency exist for all texture inputs.
 
\subsection{Shape Consistency}
The shape consistency is accomplished by generating background pixels for the locations that are outside the segmentation mask. This can be achieved by using the following loss function on the generated image $\fake = G(\col, \texture, \shape)$:
\begin{equation}
\loss_{s} = \EX_{\substack{\fake\sim P_g\\ \shape\sim p(\shape)}} \bigg[ \frac{1}{|1 - \shape|} \sum_{i,j}(1 -\shape) \cdot \Lone{\fake - \mathbf{b}} \bigg].
\label{eq:shape-consistency}
\end{equation}
Here, $1 - \shape$ is the binary complement of the input segmentation mask, $||.||_1$ is the L1 norm, and $\mathbf{b}$ is the background color, which, in our case, is white.

\vspace{-0.2cm}
\subsection{Generator Color Check}
In our experiments, we observed that the color consistency loss in equation~\ref{eq:color-consistency} sometimes causes the average color of the generated articles to collapse into a single color. Therefore, in order to avoid trivial solutions and to stabilize the training, we put an additional average color check right after the generator as follows:
\begin{equation}
\loss_{g} = \EX_{\col\sim p(\col)} \bigg[\EX_{\substack{\fake^{\col}\sim P_g^{\col}\\ \shape\sim p(\shape)}} \Big[\Ltwo{\col - A(\fake^{\col}, \shape)}\Big]\bigg].
\end{equation}

\vspace{-0.2cm}
\subsection{Generator Loss Function}
We aggregate the aforementioned losses and train our generator using the following function:
\begin{equation}
\min_{G} \loss_{W} + \loss_{\text{aux}} + \lambda_{c} \loss_{c} + \lambda_{t} \loss_{t} + \lambda_{s} \loss_{s} + \lambda_g \loss_g.
\end{equation}
Here, $\lambda_c, \lambda_t, \lambda_s, \lambda_g$ are the weights for the loss functions for color, texture, shape, and generator check, respectively.

\section{Experiments}

\subsection{Setup}
In our experiments, we used a server with Intel Xeon CPU (E5-2630), 256GB system memory, and an NVIDIA P100 GPU with 12GB of graphics memory. We modified the Tensorflow code\footnote{\urlstyle{tt}Our source code is available at \url{https://github.com/zalandoresearch/disentangling_conditional_gans}} from~\cite{progressive_gan}, where they used a temporal smoothing over the generator to create photorealistic images. Unlike that work, we did not progressively grow our model. Instead, we trained a network to directly generate $128 \times 128$ pixel images. Both our generator and discriminator layers are the same as in~\cite{progressive_gan}. Our mask embedding network has the same architecture as our discriminator, except the last layer, which outputs a 512-dimensional vector and normalizes it with the L2-norm.

We train our model using ADAM as optimizer~\cite{adam} with the following parameters: learning rate$=0.001$, $\beta_1=0$, $\beta_2=0.99$, $\epsilon = 10^{-8}$. The weights of our generator loss functions are $\lambda_c=100, \lambda_t=100, \lambda_s=1, \lambda_g=100$. 

Our training dataset is composed of over 120,000 images of dresses that are downloaded from Zalando's website\footnote{\url{www.zalando.de}}. These images are padded into squares of size $128 \times 128$ pixels. At each training iteration, we sample two random colors, textures, and masks, and feed them into our generator using all combinations, which gives us a batch size of $2^3=8$. This combinatorial batch, with the help of our loss functions, disentangles the effects of input attributes. This is because we check if the effect of a certain attribute stay roughly the same for each combination.

In our experiments, for input and output images, we used the standard RGB color space representation. One can easily use CIE-LAB color space with small modifications to our architecture.

\vspace{-0.2cm}
\subsection{Results: Color Control}
\begin{figure}[h]
	\centering
	\includegraphics[width=0.45\textwidth]{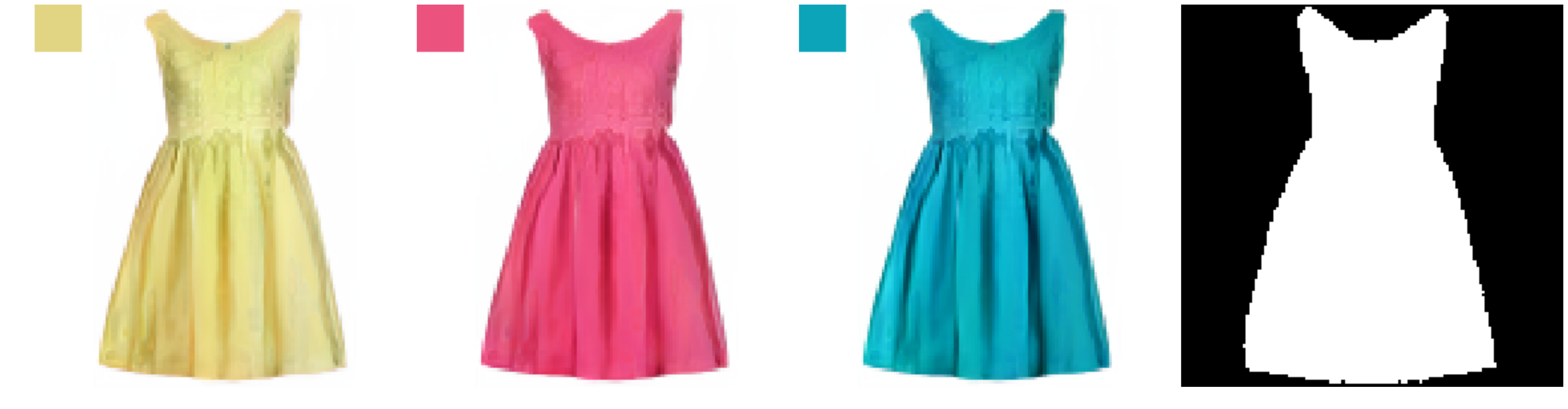}
	\caption{Generated images using randomly-chosen colors (represented on the top left) and constant texture and shape.}
	\label{fig:color-control}
\end{figure}

We demonstrate the effect of color tuning in Figure~\ref{fig:color-control}, where we randomly select three colors (represented as a colored square at the top-left corner) and generate three articles using the same texture and shape. We observe that the average colors of the generated articles are approximately the same with their corresponding input colors. In addition, the local structure and the shape are very stable and mostly unaffected by the input color changes.

\begin{figure}[h]
	\centering
	\includegraphics[width=0.45\textwidth]{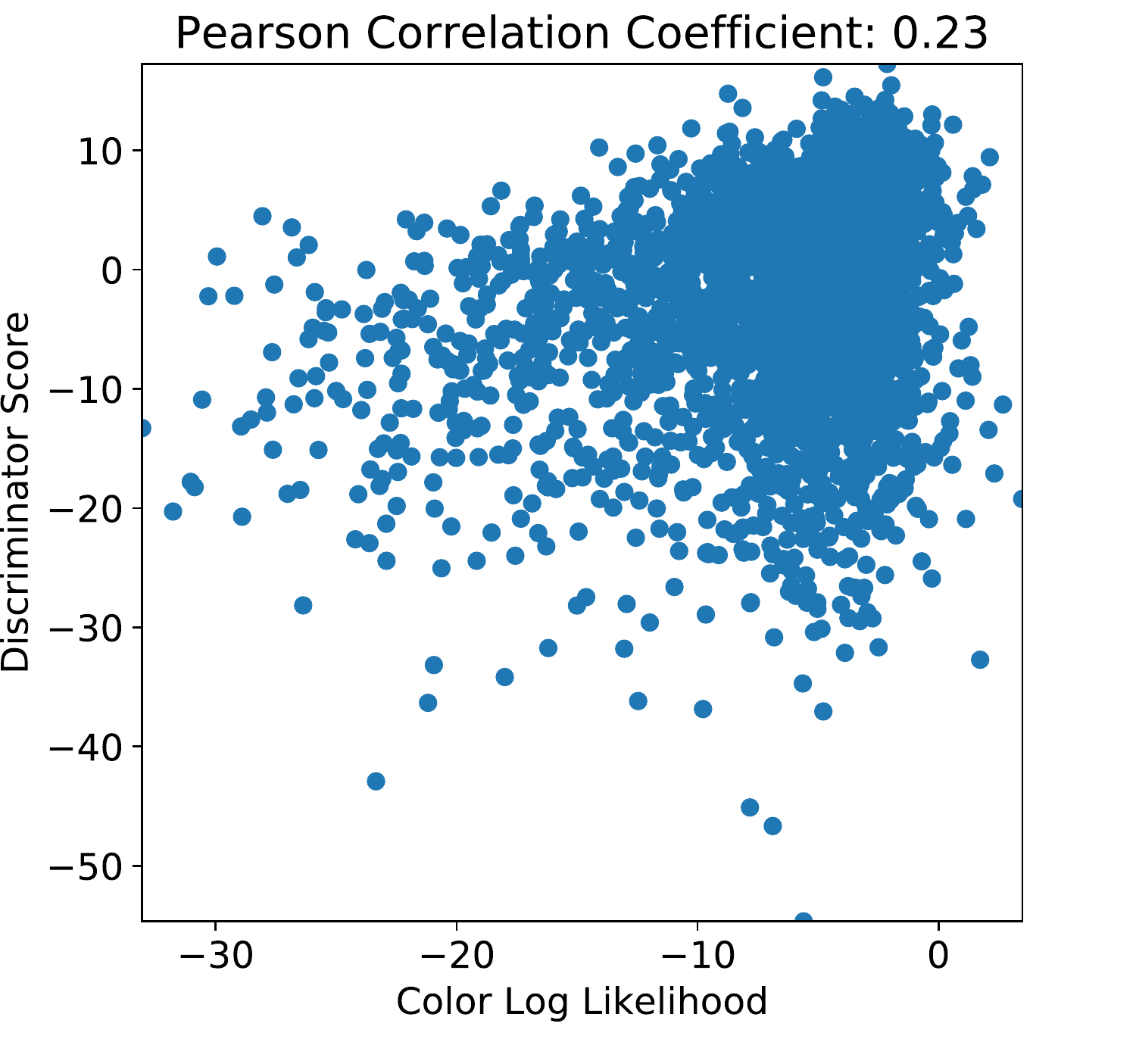}
	\vspace{-0.5cm}
	\caption{The log-likelihood of randomly-chosen input colors and discriminator scores of images with these colors.}
	\label{fig:color-likelihood}
\end{figure}

The color distribution of real articles is not uniform. This can limit the generator output to a certain set of colors, as the discriminator might reject a plausible looking dress with an unlikely color. In Figure~\ref{fig:color-likelihood}, for around 3000 generated images, we plot the log-likelihood of their input colors against their discriminator scores. We measure the color likelihood by fitting a Gaussian Mixture Model (16 components) on the colors of real articles. We can observe that, although there is a small linear correlation, our discriminator is not strongly biased towards colors of the real articles. This is especially important when designers want to create novel garments.
\vspace{-0.2cm}

\subsection{Results: Texture Control}
\begin{figure}[h]
	\centering
	\includegraphics[width=0.45\textwidth]{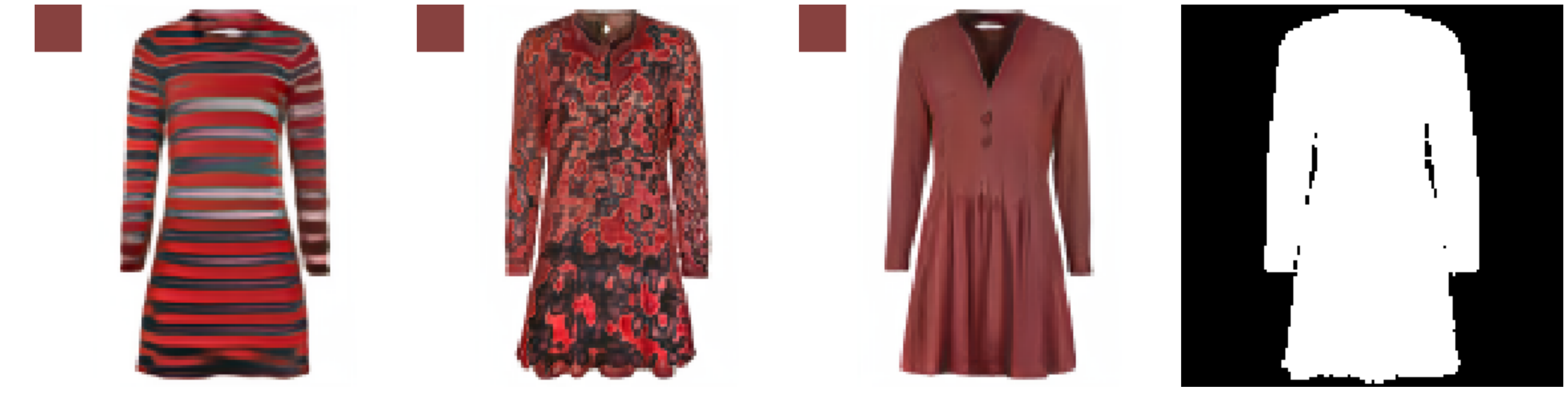}
	\caption{Generated images using randomly-chosen textures with constant color and shape inputs.}
	\label{fig:texture-control}
\end{figure}

In order to show the texture control, we randomly sample three latent vectors and generate images using the same color and shape. We can see in Figure~\ref{fig:texture-control} that different texture inputs create distinct local structures and patterns, yet the average color and the shape of the articles are equivalent.

\subsection{Results: Shape Control}
\begin{figure}[h]
	\centering
	\includegraphics[width=0.45\textwidth]{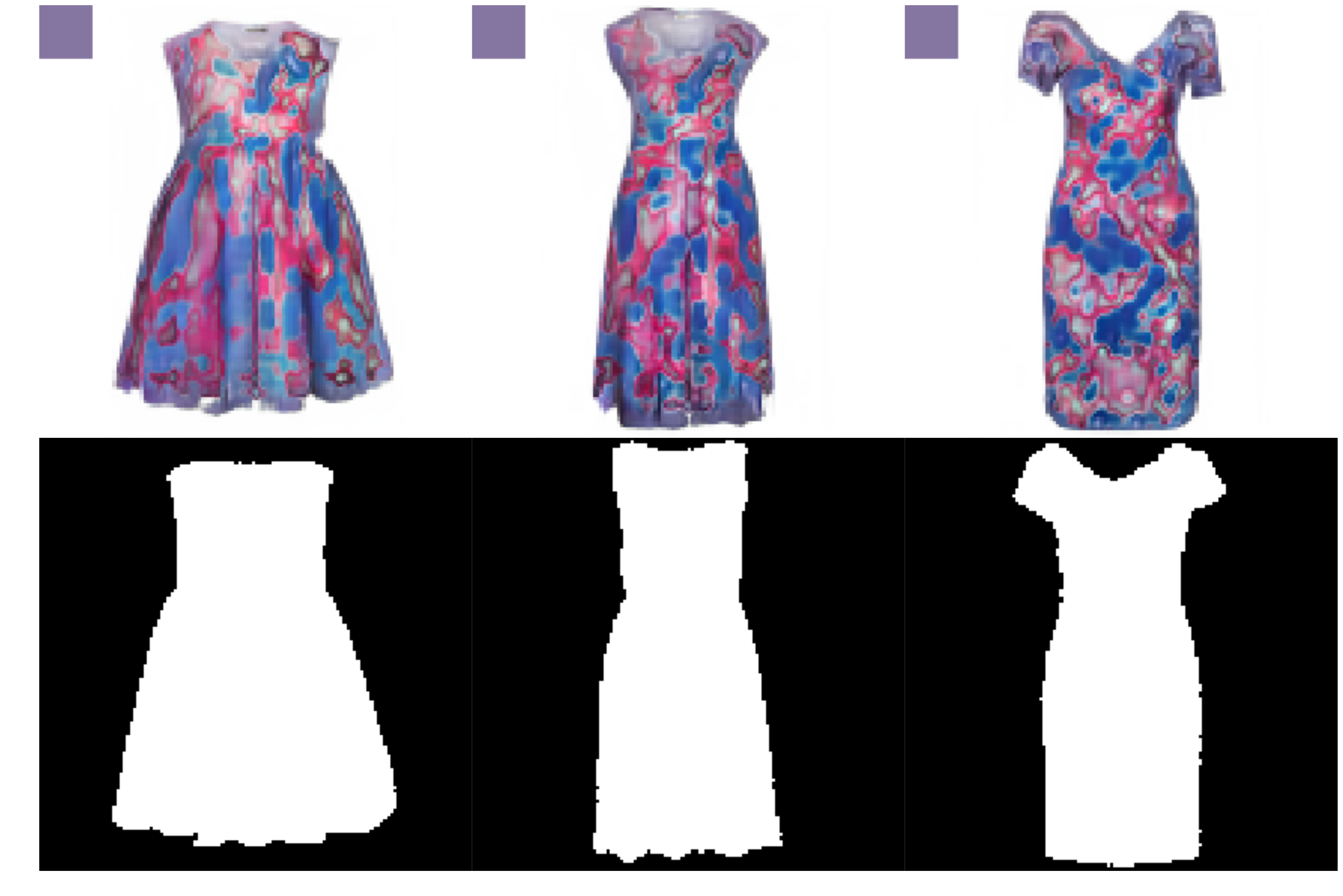}
	\caption{Generated images using randomly chosen shapes with constant color and texture.}
	\label{fig:shape-control}
\end{figure}

In our generator, the shape of a generated article is guided by providing a binary mask. Similar to color and texture, we want to adjust the shape of the generated article independent of the color and texture inputs. In Figure~\ref{fig:shape-control}, we demonstrate three synthesized images that have the same color and texture, but different shapes. We observe that, although not pixel precise, the garment outlines faithfully follow the input shape mask, which is sufficient for design visualizations. One can see that the local structure of an article cannot stay perfectly unchanged, when we change its shape. However, our generator creates realistic article images with different shapes yet similar-looking textures. These images are superior to cut-out versions of 2D pattern maps.

\begin{figure}[h]
	\centering
	\includegraphics[width=0.45\textwidth]{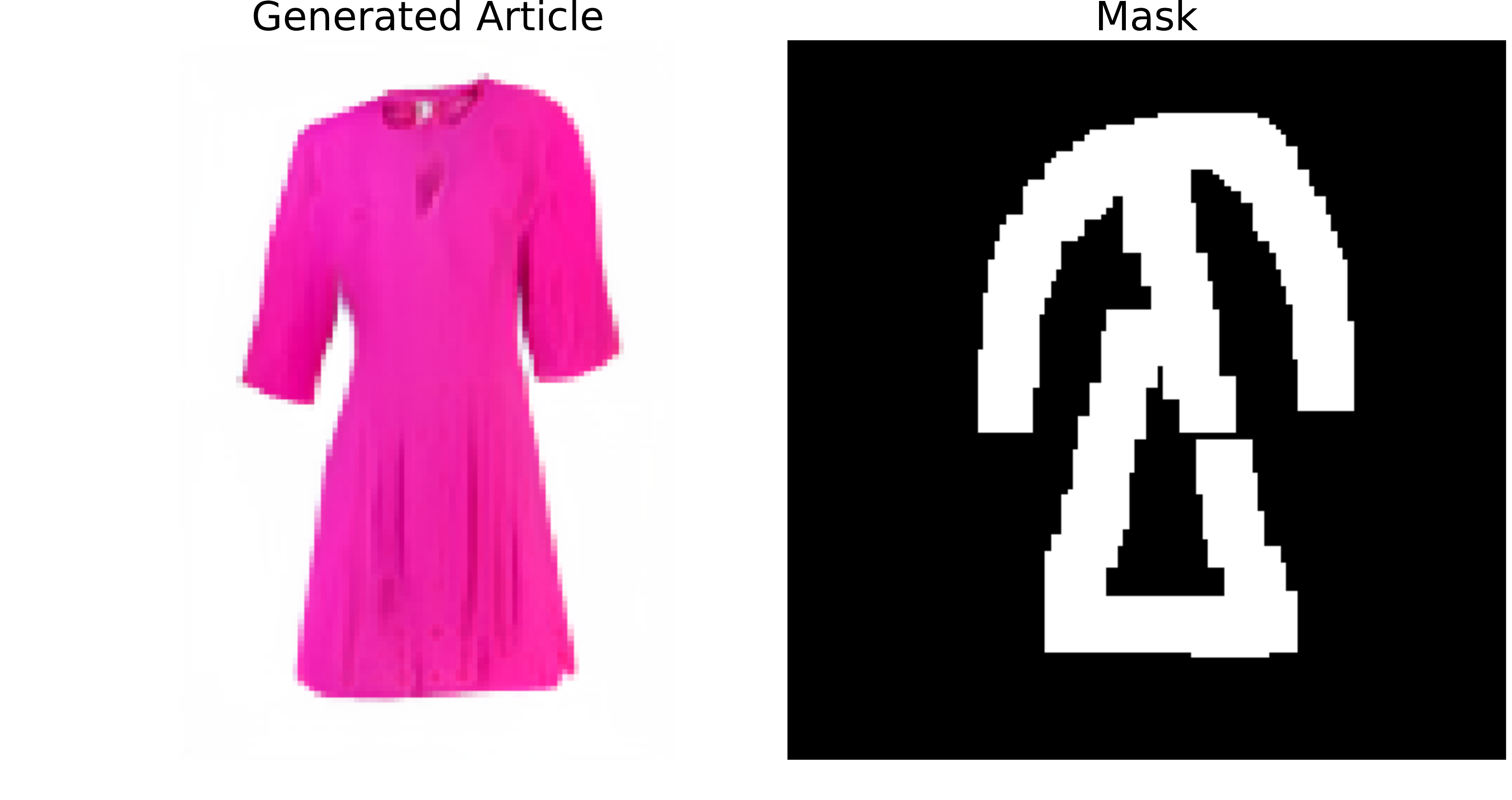}
	\caption{An image generated with a hand-drawn mask.}
	\label{fig:custom-mask}
\end{figure}

Next, we investigate the stability of the generated images, when we input masks that do not come from real mask distribution. In Figure~\ref{fig:custom-mask}, we can see that, although the mask is hand drawn and unlikely to belong to a real article, its corresponding article looks plausible. This property can be used in a fashion design user interface to define article outlines.

\section{Fashion Design}
\begin{figure}[h]
\subfloat[][Real article]{\includegraphics[width=0.16\textwidth]{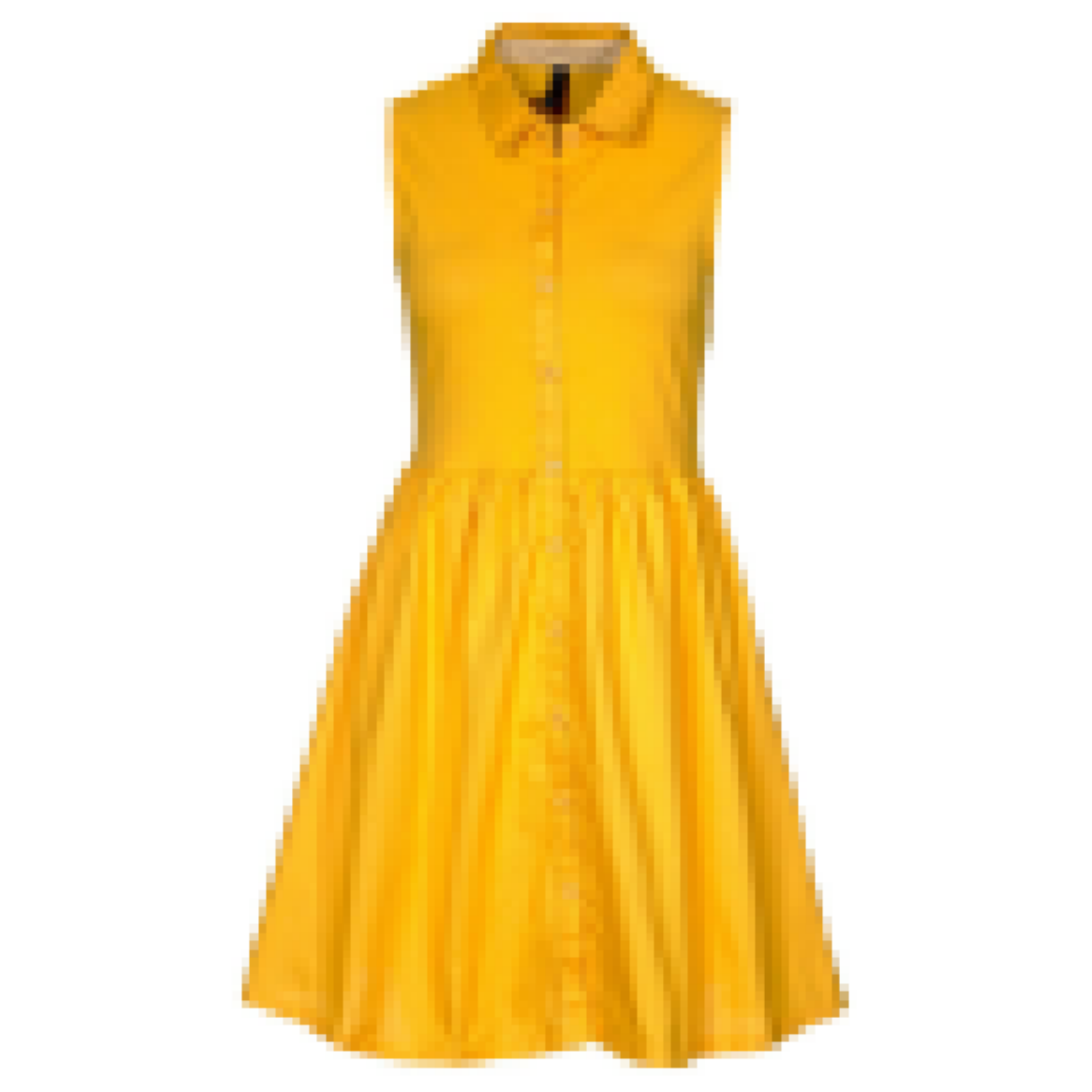}}
\subfloat[][Reconstructed article]{\includegraphics[width=0.16\textwidth]{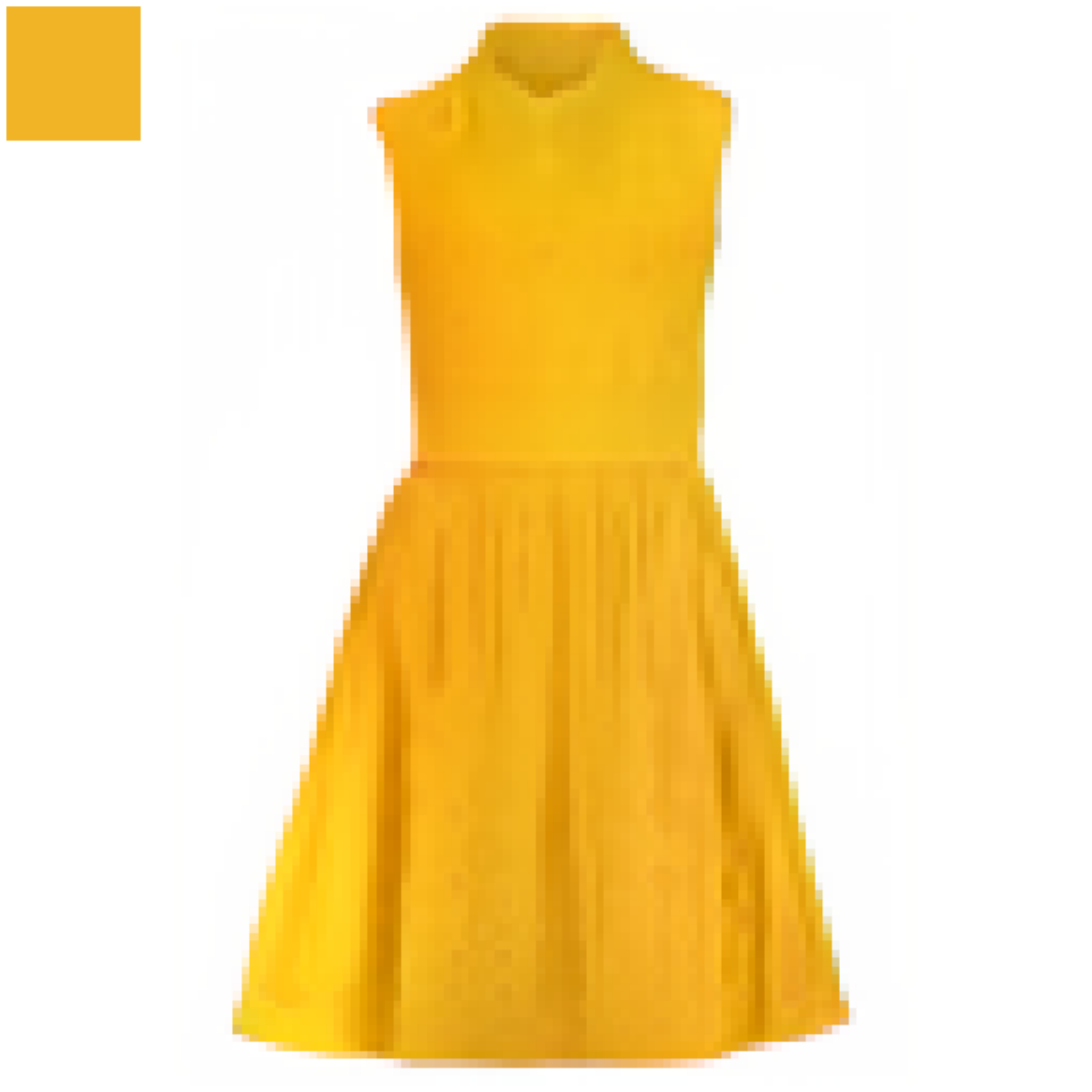}}
\subfloat[][Shape mask]{\includegraphics[width=0.16\textwidth]{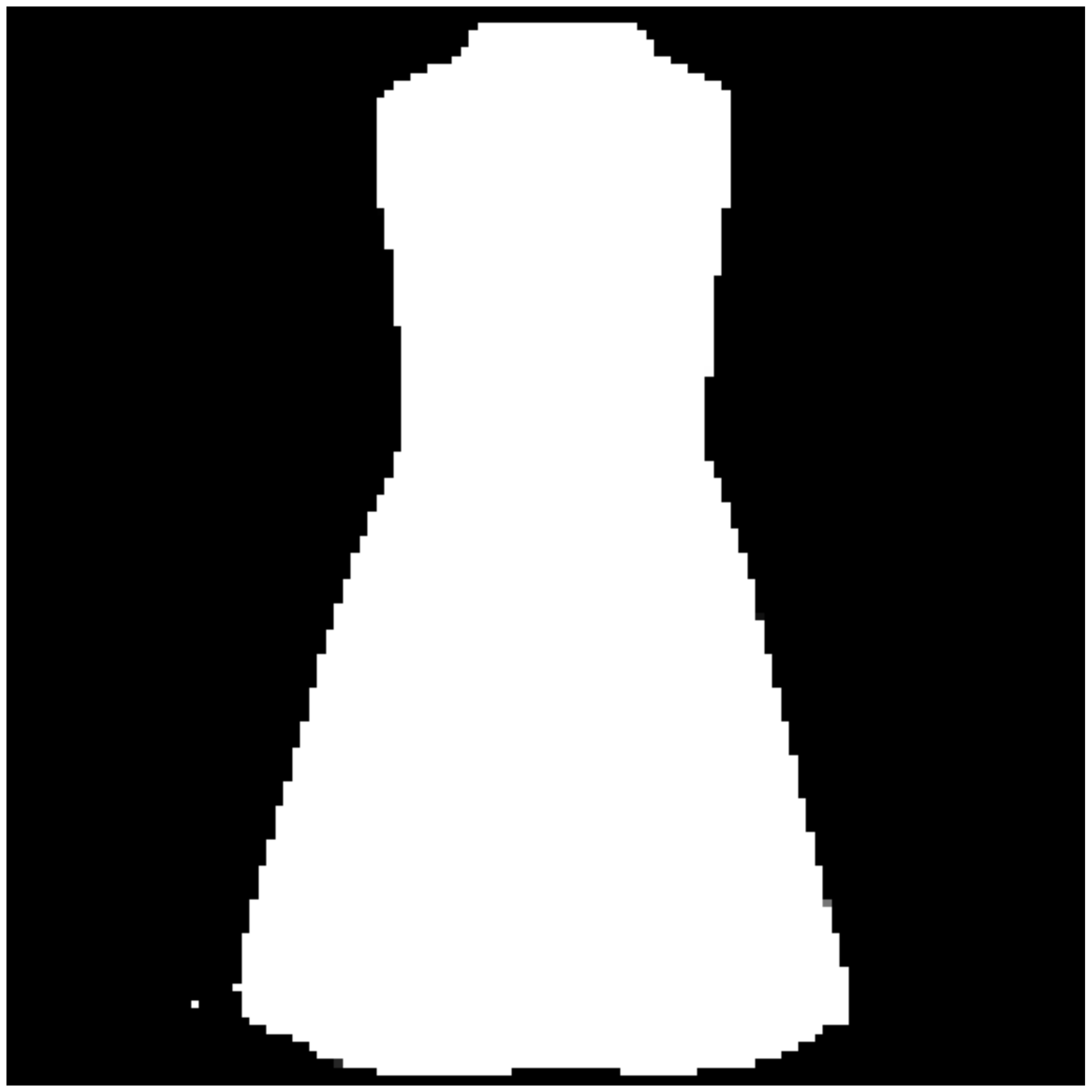}}
\caption{Reconstruction of a real article using our generator. Estimated color is represented at the top left.}
\label{fig:real-recon}
\end{figure}

In order to modify images of real fashion articles, we need to find their corresponding color, texture, and shape inputs. We calculate the shape (or segmentation mask) of a garment by using a simple neural network that is trained on article images and their corresponding binary masks. One can obtain the mask of an image by using an interactive method such as GrabCut~\cite{grabcut}. Once we have the shape estimate $\hat{\shape}$, we can estimate the average color $\hat{\col}$ using equation~\ref{eq:average-color}. The texture (or the local structure) of an article can be obtained by solving the following optimization problem:
\begin{equation}
\hat{\texture}= \argmin_{\texture} \Lone{\real - \fake^{\texture}} + \beta_t \loss_t\big(\real, \fake^{\texture} \big) + \beta_{\text{KL}} \text{KL}\big(\mathcal{N}(0, 1), \texture \big),
\label{eq:optimization}
\end{equation}
where, $\fake^{\texture} = G(\hat{\col}, \texture, \hat{\shape})$, $\loss_t(\cdot,\cdot)$ is the texture consistency loss, and $\text{KL}(\cdot,\cdot)$ computes the KL-divergence between the texture vector and a zero-mean, unit-variance normal distribution as follows:
\begin{equation}
\text{KL}\big(N(0, 1), \mathcal{N}(\mu_{\texture}, \sigma_{\texture})\big) = \log (\sigma_{\texture}) + \frac{1 + \mu_{\texture}^2}{2\sigma_{\texture}^2} - \frac{1}{2}.
\end{equation}
Here, $\mu_{\texture}$ and $\sigma_{\texture}$ are the mean and variance values that are calculated over the individual elements of texture vector $\texture$. KL-divergence regularizes the estimated $\hat{\texture}$ to be close to the distribution $p(\texture)$ used during training of the GAN. The weights we use during the optimization are $\beta_t=1, \beta_{KL}=0.1$. In Figure~\ref{fig:real-recon}, we demonstrate a real article and its reconstructed version using the estimated input tuple $(\hat{\col}, \hat{\texture}, \hat{\shape})$. We can see that the color and  shape inputs are accurately reflected in the reconstructed version. In addition, the estimated texture is able to capture the horizontal line in the middle, and the shading/wrinkling in the lower part of the garment.

\begin{figure}[h]
	\newcommand{\bla}{0.12}
	\subfloat[][Reconstructed]{\includegraphics[width=\bla\textwidth]{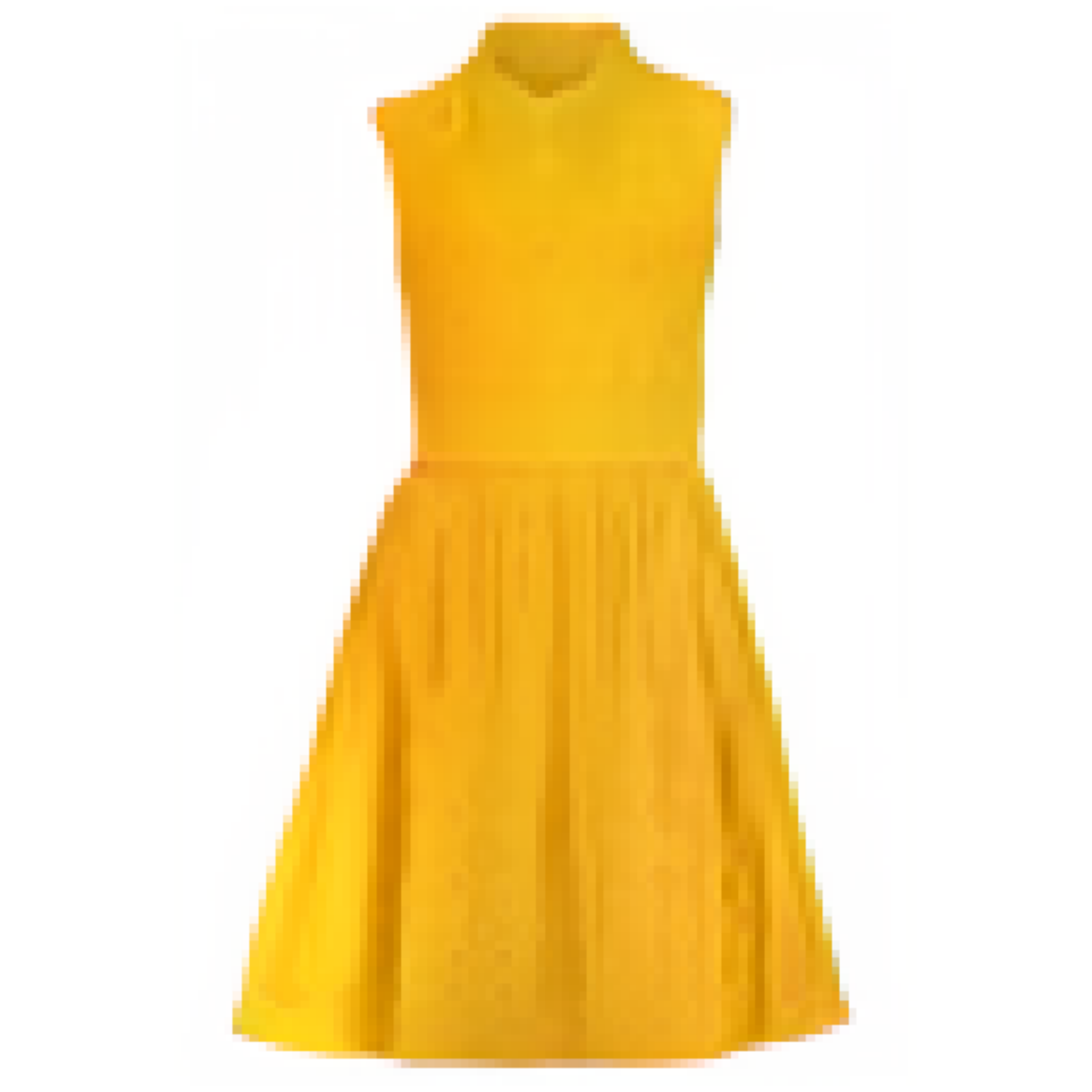}}
	\subfloat[][Shape]{\includegraphics[width=\bla\textwidth]{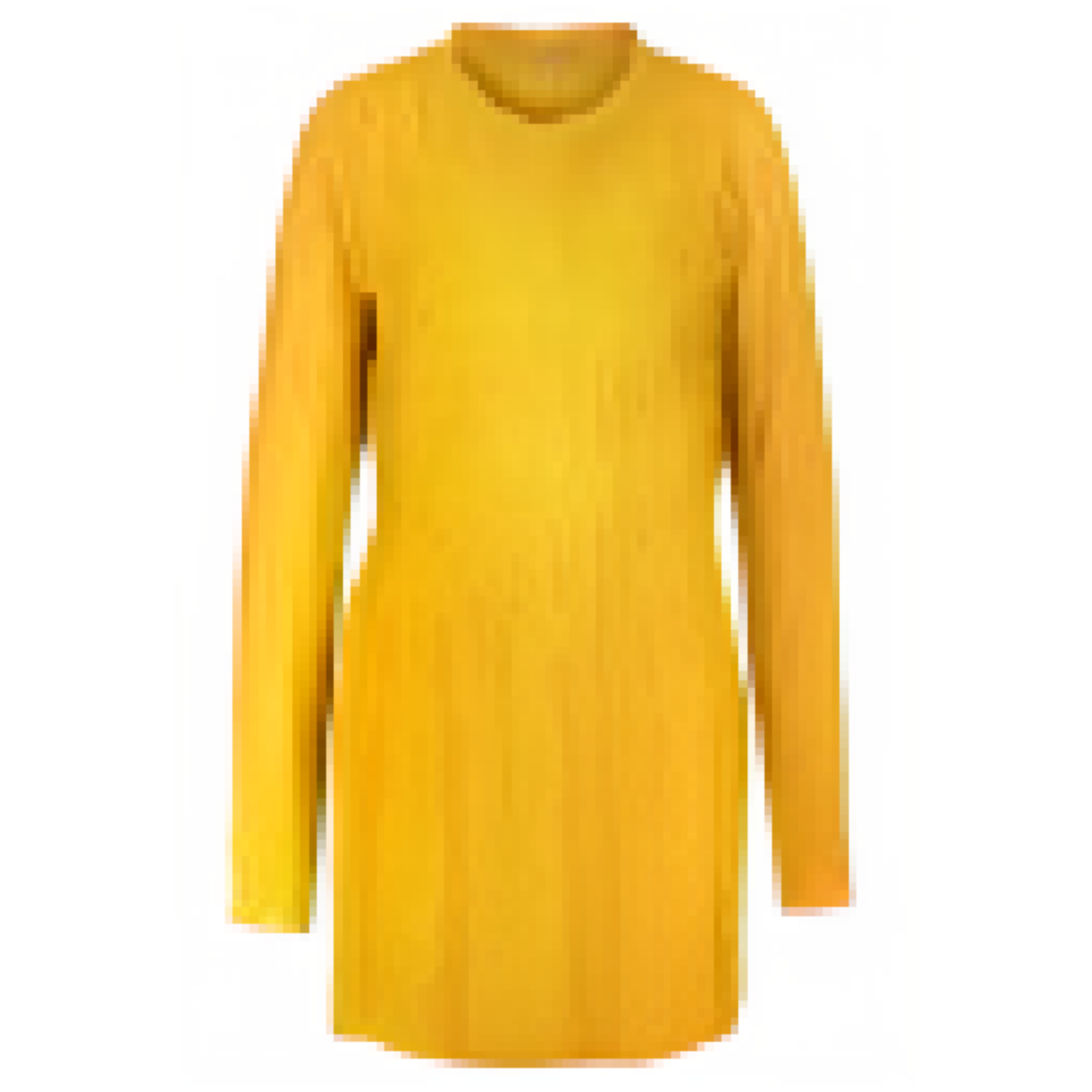}}
	\subfloat[][Texture]{\includegraphics[width=\bla\textwidth]{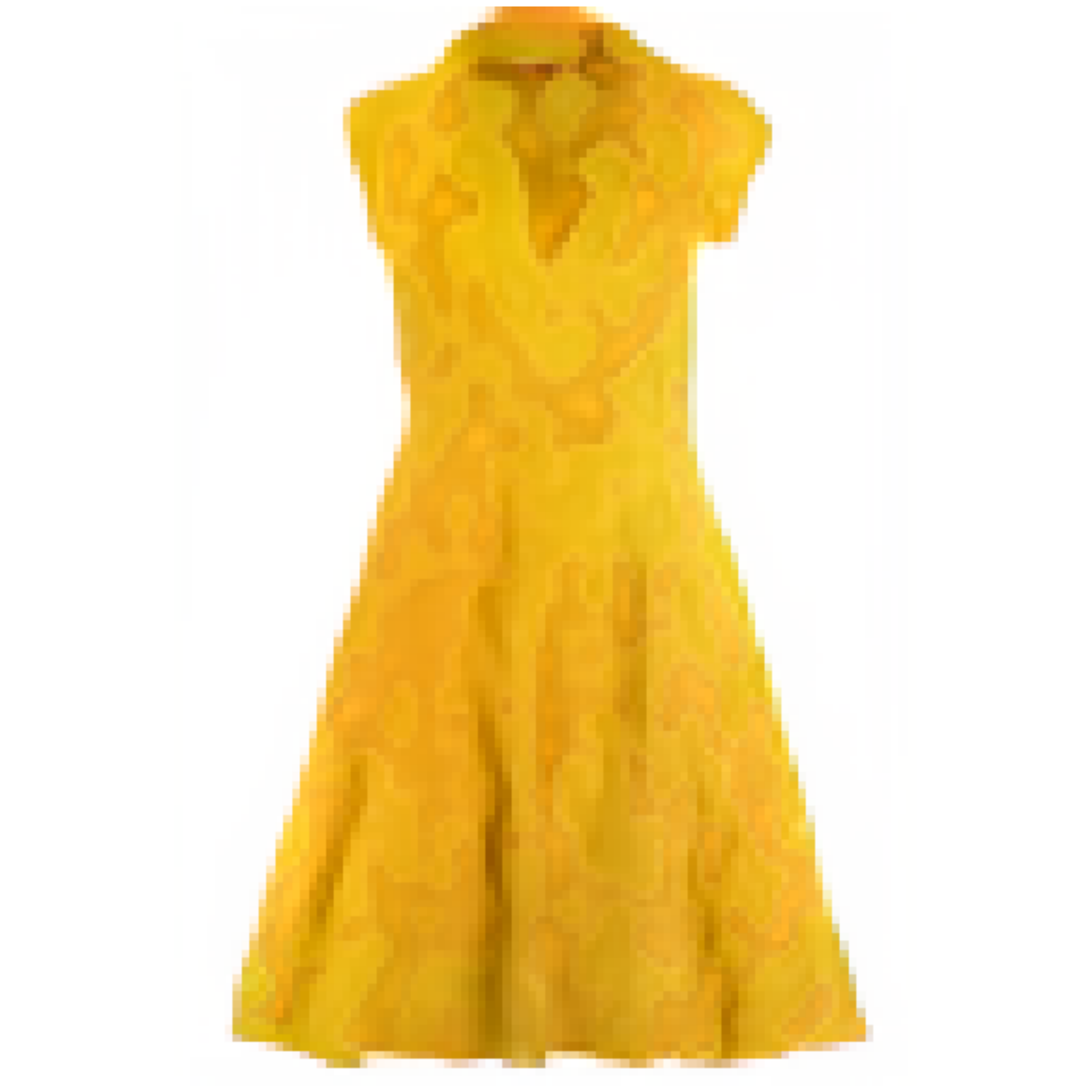}}
	\subfloat[][Texture+Shape]{\includegraphics[width=\bla\textwidth]{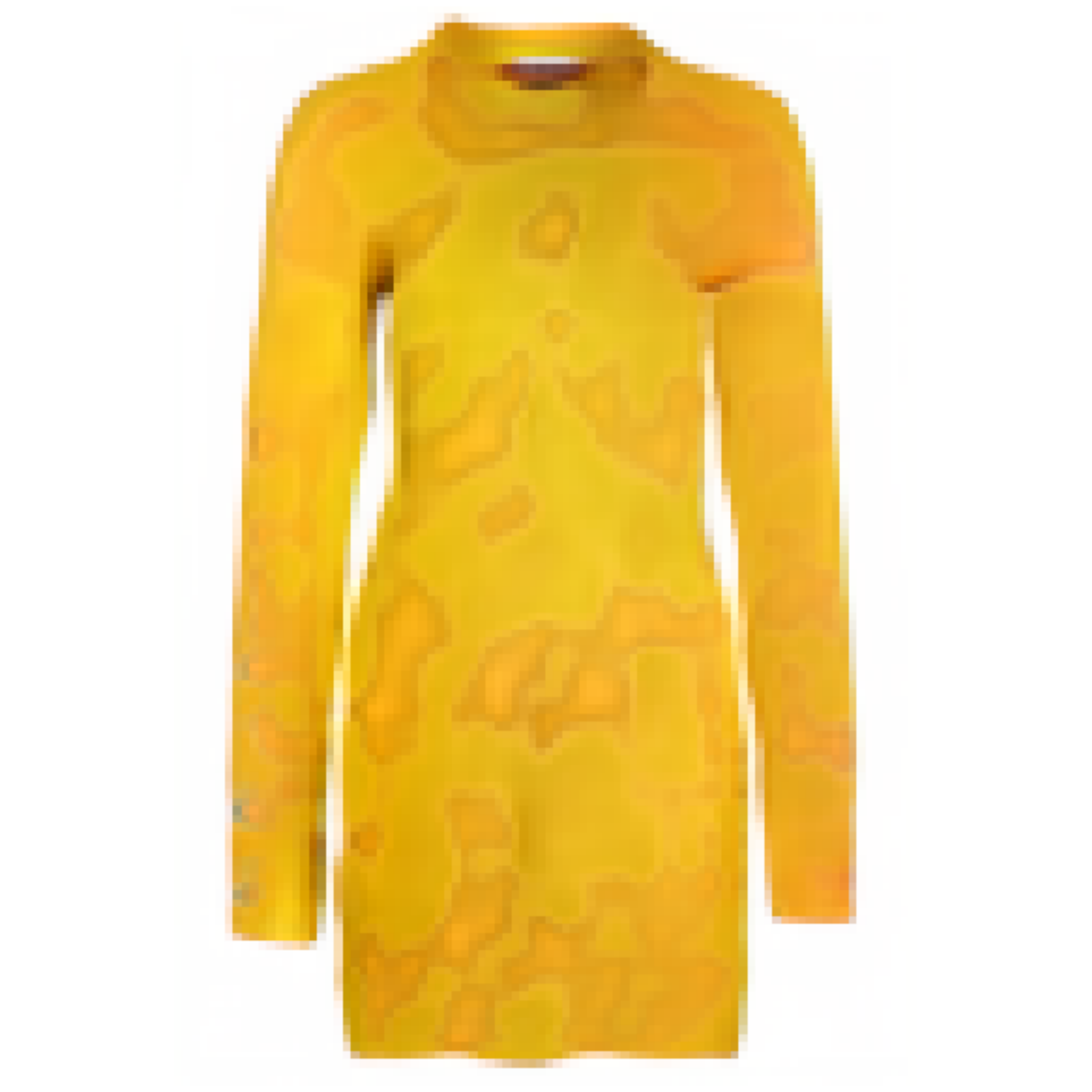}}\\
	\subfloat[][Color]{\includegraphics[width=\bla\textwidth]{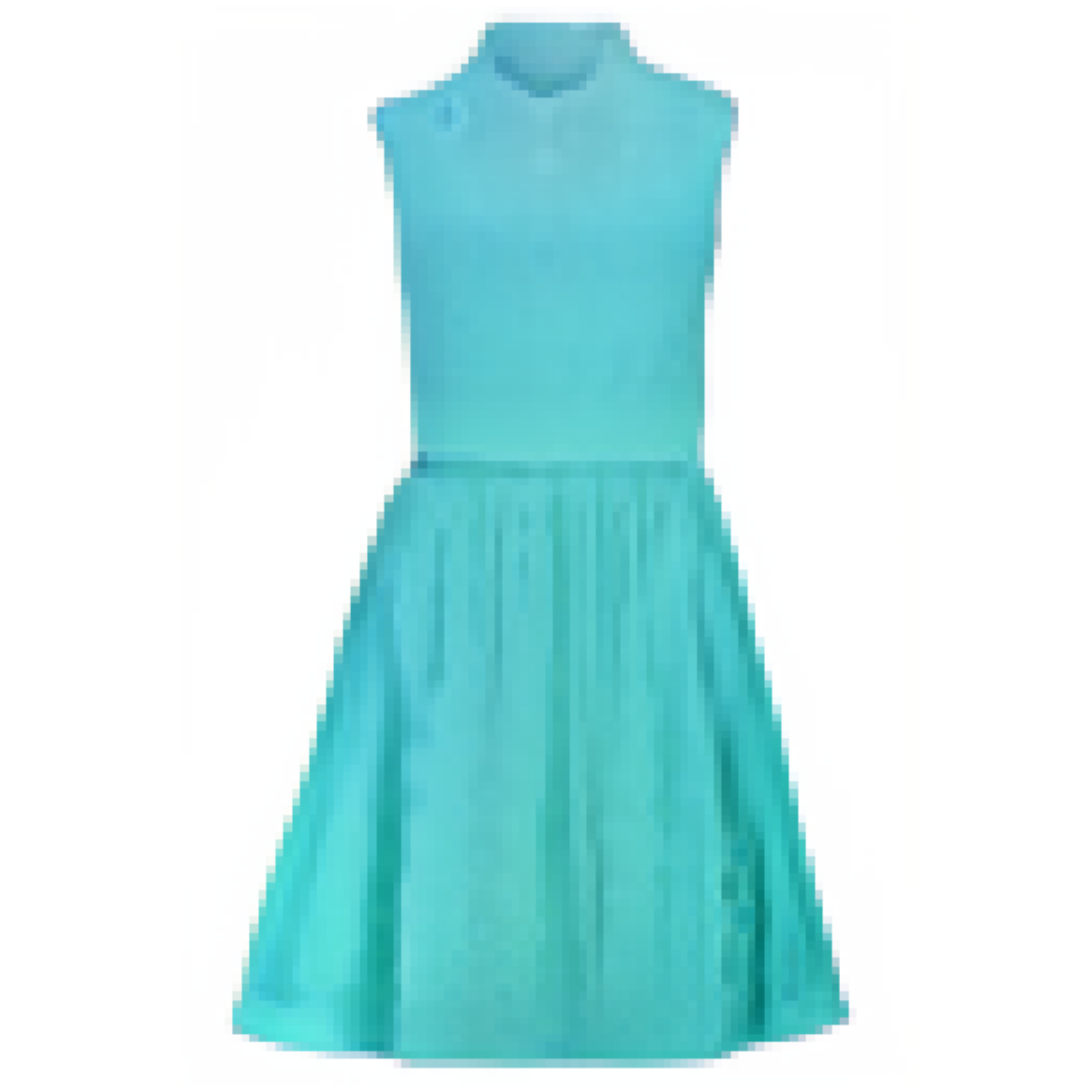}}
	\subfloat[][Color+Shape]{\includegraphics[width=\bla\textwidth]{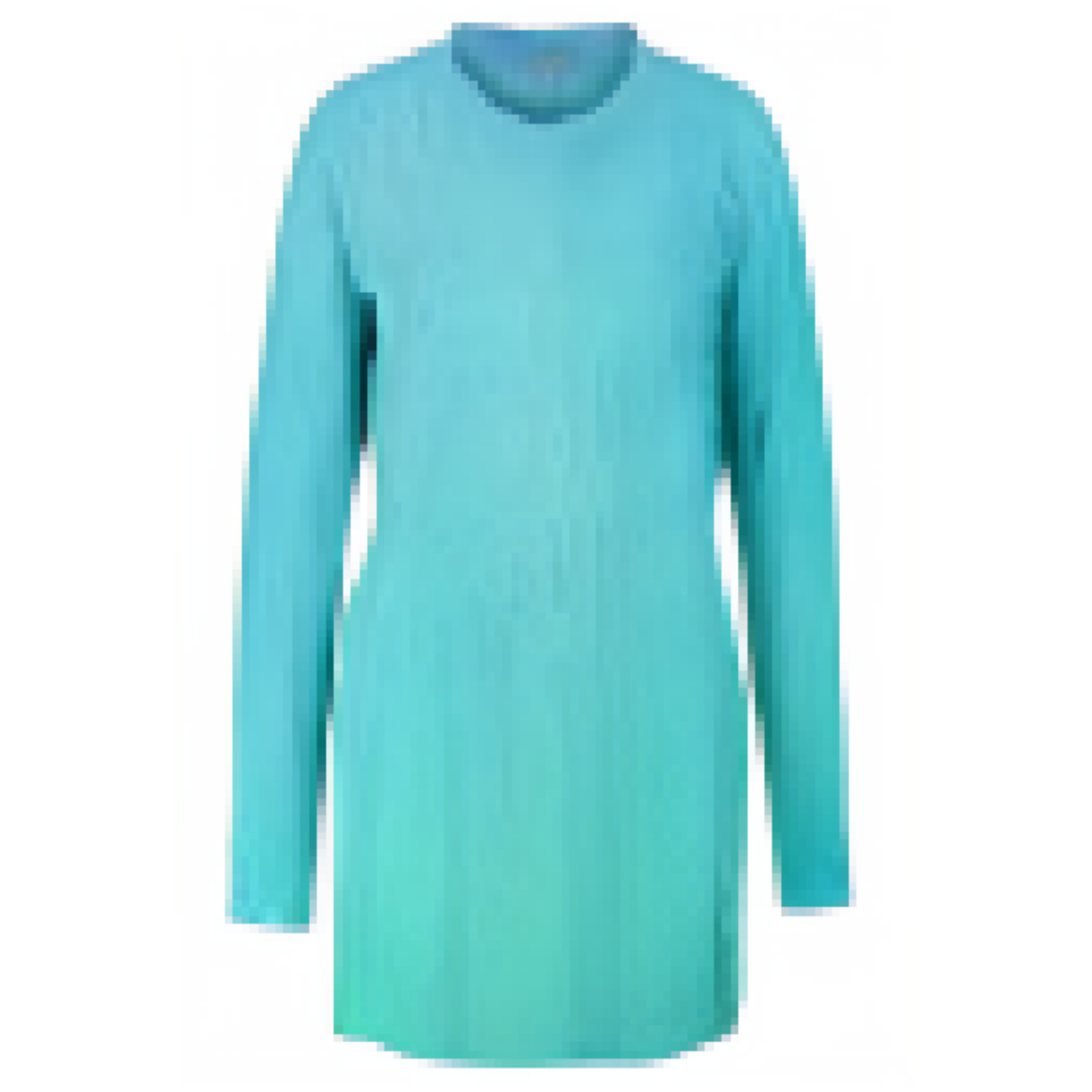}}
	\subfloat[][Color+Texture]{\includegraphics[width=\bla\textwidth]{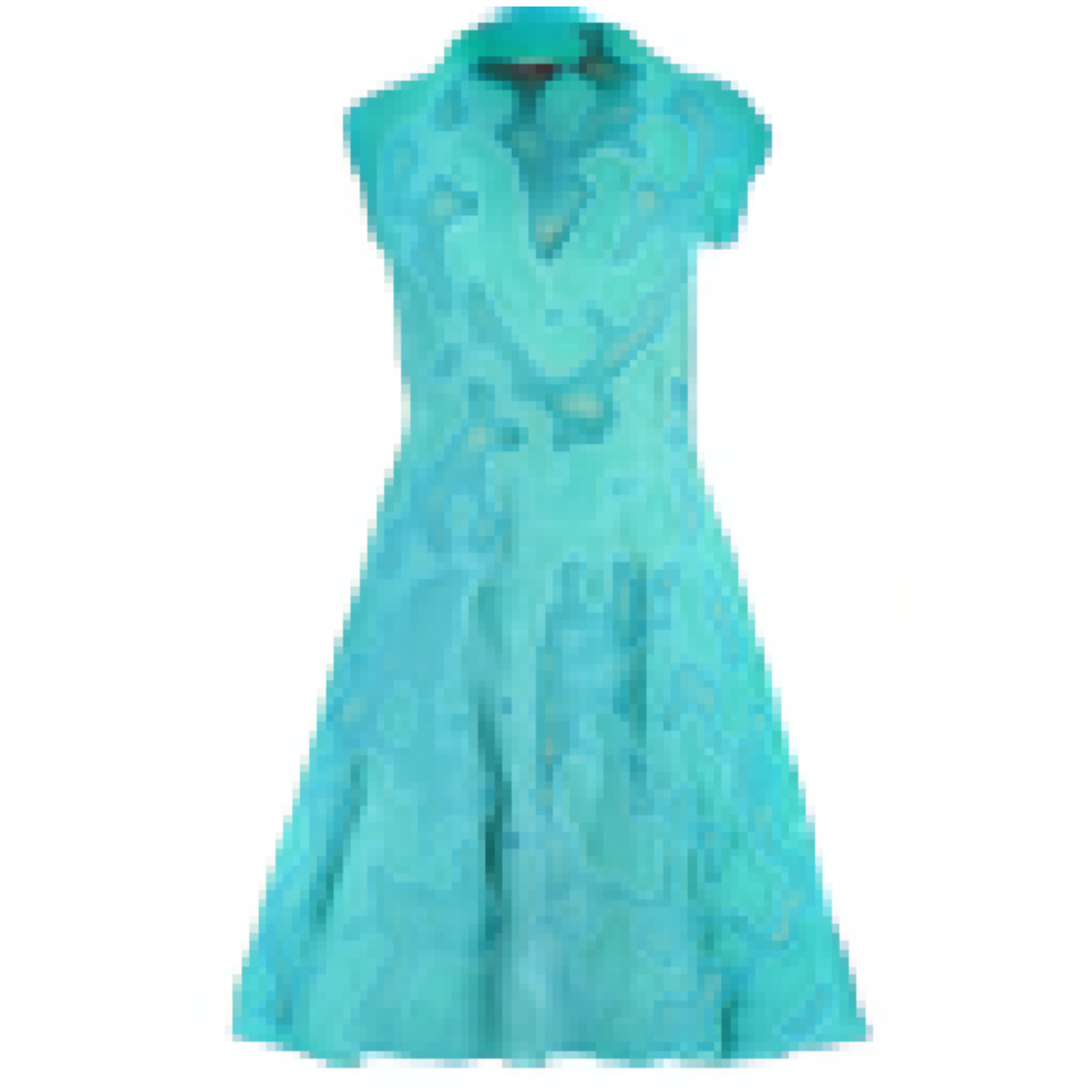}}
	\subfloat[][Color+Texture+\\Shape]{\includegraphics[width=\bla\textwidth]{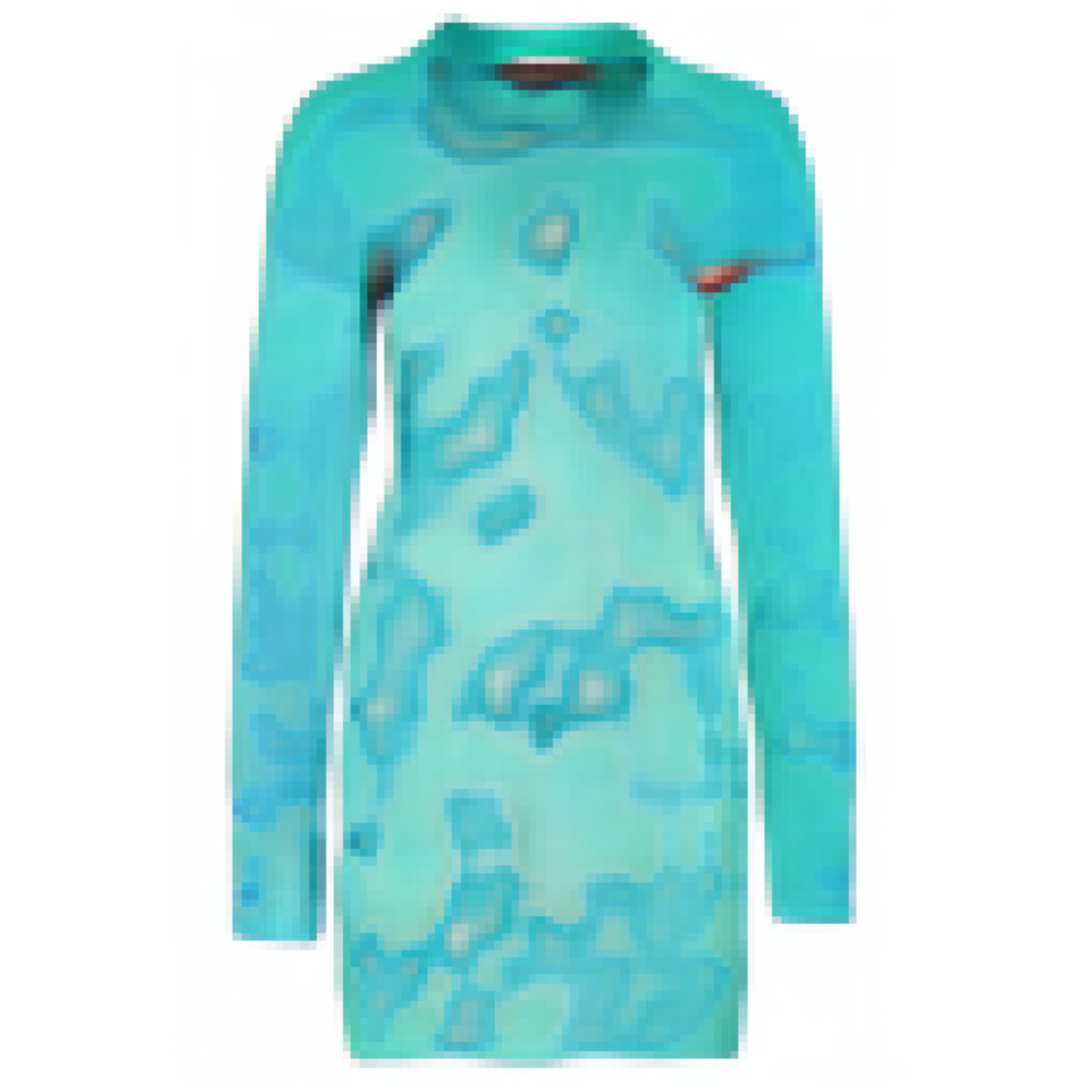}}
	\caption{An example design flow that changes input attributes of the real article in Figure~\ref{fig:real-recon}.}
	\label{fig:fashion-design}
\end{figure}

In Figure~\ref{fig:fashion-design}, we modify the input attributes to the generator and observe the process is robust and each attribute separately affects the reconstructed article.
\vspace{-0.5cm}

\vspace{-0.5cm}
\subsection{Discussions}
\begin{figure}[h]
	\subfloat[][Real article]{\includegraphics[width=0.16\textwidth]{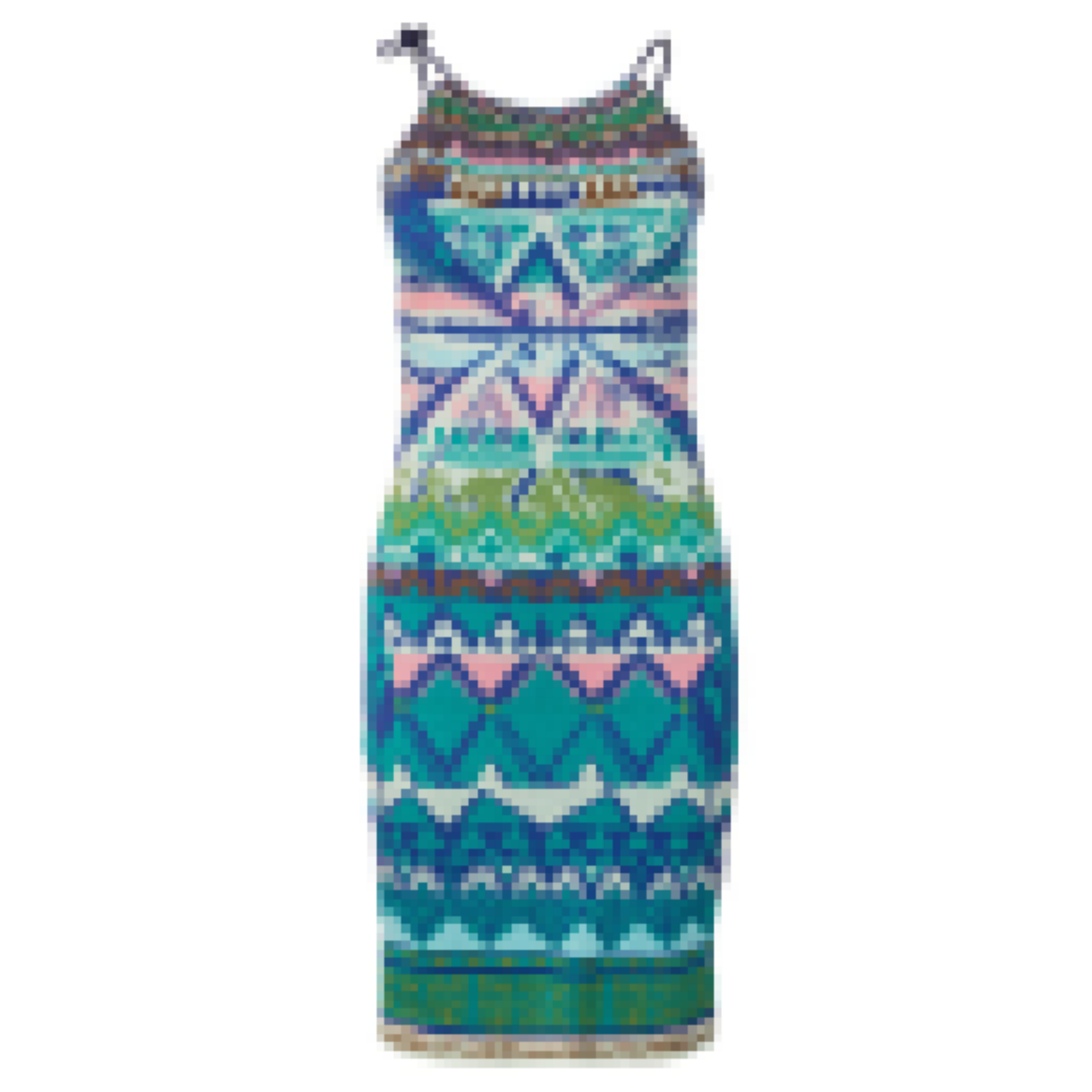}}
	\subfloat[][Reconstructed article]{\includegraphics[width=0.16\textwidth]{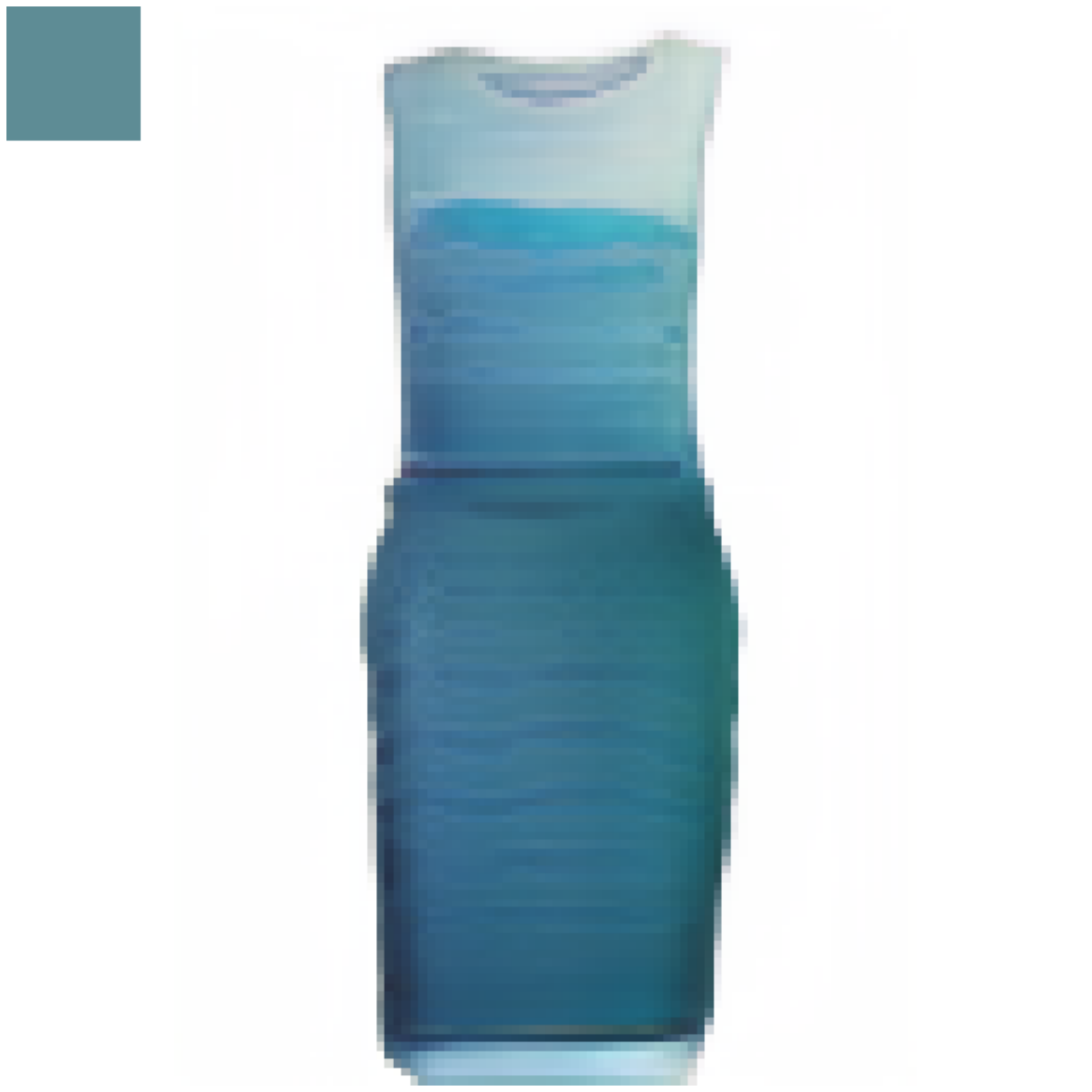}}
	\subfloat[][Shape mask $\hat{\shape}$]{\includegraphics[width=0.16\textwidth]{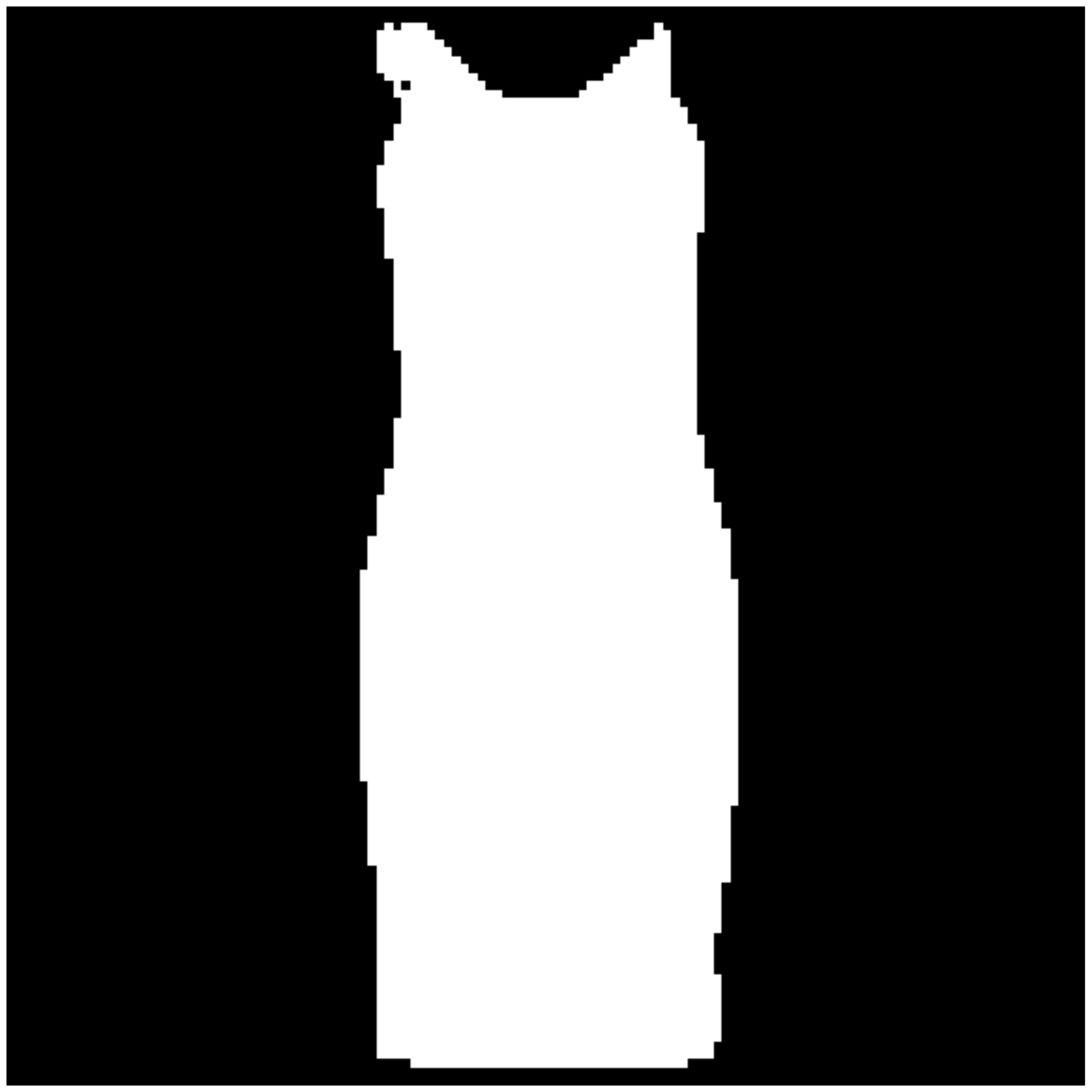}}
	\caption{Failed reconstruction due to multi-colored and highly-textured article.}
	\label{fig:real-recon-fail}
\end{figure}

Our generator only ensures the average color within the mask will be the same as the input color. This assumption limits the generator and prevents it from capturing some parts of the real image distribution $P_r$. For example, the multi-colored garment in Figure~\ref{fig:real-recon-fail} is not accurately reconstructed, due to its complex structure. 
\vspace{-0.5cm}

It is possible to extend the color input from a single color to a collection of colors or to a color histogram. In this case, however, a single shape mask might not be enough to check if the colors are correctly generated at correct image locations. Instead, a full color segmentation of the image might be required.

The Laplacian matting matrix in our texture consistency loss is computed using a $3 \times 3$ neighborhood around each pixel. If we increase the neighborhood size, we can capture larger structures on the articles without changing the size of the Laplacian matting matrix. However, the sparsity of this matrix decreases quadratically with the neighborhood size, which would dramatically increase the computation time and memory requirements.

One can use pix2pix~\cite{image_to_image} between shape masks and real images (color and texture as conditional inputs). However, as stated in~\cite{image_to_image}, these architectures tend to ignore the noise inputs and learn a deterministic mapping, which is not desirable in a design system.

\section{Conclusions}
In this paper, we presented a generative adversarial network architecture and a corresponding training procedure that takes color, texture, and binary shape mask as input, and outputs an image of a fashion article. We showed that by using our consistency loss functions, we were able to disentangle the effects of generator inputs, which enabled us to independently tune the attributes of a generated image. Our generator presents an opportunity to easily design and modify fashion images.

The attributes we presented here are only a subset of characteristics that fashion designers require. We plan to add more sophisticated control over the generation process by extending our method to multiple color inputs and allow texture input directly from an image or another article.

\bibliographystyle{ACM-Reference-Format}
\bibliography{sample-bibliography}

\end{document}